\documentclass{article}

\usepackage[dvipsnames]{xcolor}
\usepackage{bm}
\usepackage{xspace}
\newcommand{\boldPara}[1]{\vspace{0.0in}\noindent \textbf{#1}}
\newcommand\cmt[1]{\vspace{0.05in}\noindent\textcolor{OliveGreen}{.: #1 :.}}
\newcommand\td[1]{\textcolor{NavyBlue}{\textbf{Todo.} #1}}

\renewcommand\td[1]{}
\renewcommand\cmt[1]{}
\newcommand\acmt[1]{\textcolor{Purple}{\footnotesize{\texttt{// #1}}}}

\makeatletter
\newcommand{\customlabel}[2]{%
   \protected@write \@auxout {}{\string \newlabel {#1}{{#2}{}{}{}{}} }%
}
\makeatother

\usepackage{microtype}
\usepackage{wrapfig}
\usepackage{float}
\usepackage[pdftex]{graphicx}
\usepackage[T1]{fontenc}    
\usepackage{amsfonts}       
\usepackage{nicefrac}       
\usepackage{subcaption}
\usepackage{tikz}
\usepackage{textcomp}

\usepackage{array}
\newcolumntype{P}[1]{>{\centering\arraybackslash}p{#1}}
\newcolumntype{M}[1]{>{\arraybackslash}m{#1}}

\usepackage[vlined, boxed, ruled, algo2e]{algorithm2e}
\usepackage{balance}
\usepackage{enumitem}
\SetKwComment{Comment}{//}{}
\usepackage{flushend}
\usepackage{hyperref}
\usepackage{amsfonts}       
\usepackage{nicefrac}       
\usepackage{balance}

\usepackage[accepted]{./style/sysml2019}

\sysmltitlerunning{MotherNets: Rapid Deep Ensemble Learning}

\begin{document}

\twocolumn[
\sysmltitle{MotherNets: Rapid Deep Ensemble Learning}
\sysmlsetsymbol{equal}{*}

\begin{sysmlauthorlist}
\sysmlauthor{Abdul Wasay}{to}
\sysmlauthor{Brian Hentschel}{to}
\sysmlauthor{Yuze Liao}{to}
\sysmlauthor{Sanyuan Chen}{to}
\sysmlauthor{Stratos Idreos}{to}
\end{sysmlauthorlist}

\sysmlcorrespondingauthor{Abdul Wasay}{awasay@seas.harvard.edu}

\sysmlaffiliation{to}{Harvard School of Engineering and Applied Sciences}

\sysmlkeywords{}

\vskip 0.2in

\begin{abstract}
Ensembles of deep neural networks significantly improve 
generalization accuracy. 
However, training neural network ensembles requires a large 
amount of computational resources and time.  
State-of-the-art approaches either train all networks from 
scratch leading to prohibitive training cost that allows 
only very small ensemble sizes in practice, or generate ensembles 
by training a monolithic architecture, which results 
in lower model diversity and decreased prediction accuracy.
We propose MotherNets to enable higher accuracy and practical 
training cost for large and diverse neural network ensembles: 
A MotherNet captures the structural similarity across some 
or all members of a deep neural network ensemble which allows 
us to share data movement and computation costs across these 
networks. We first train a single or a small set of MotherNets 
and, subsequently, we generate the target ensemble networks 
by transferring the function from the trained MotherNet(s). Then, we 
continue to train these ensemble networks, which now converge 
drastically faster compared to training from scratch.
MotherNets handle ensembles with diverse architectures 
by clustering ensemble networks of similar architecture 
and training a separate MotherNet for every cluster. 
MotherNets also use clustering to control the accuracy vs. 
training cost tradeoff. 
We show that compared to state-of-the-art approaches such as 
Snapshot Ensembles, Knowledge Distillation, and TreeNets, 
MotherNets provide a new Pareto frontier for the accuracy-training 
cost tradeoff. 
Crucially, training cost and accuracy improvements continue 
to scale as we increase 
the ensemble size (2 to 3 percent reduced absolute test 
error rate and up to 35 percent faster training compared 
to Snapshot Ensembles).
We verify these benefits over numerous neural network architectures 
and large data sets. 
\end{abstract}
]

\printAffiliationsAndNotice{}  


\def\checkmark{\tikz\fill[scale=0.4](0,.35) -- (.25,0) -- (1,.7) -- (.25,.15) -- cycle;}
\def\checkmarkb{\tikz\fill[scale=0.45](0,.35) -- (.25,0) -- (1,.7) -- (.25,.15) -- cycle;}

\section{Introduction}

\boldPara{Neural network ensembles.} 
Various applications increasingly use ensembles of multiple 
neural networks to scale the representational power of their 
deep learning pipelines. 
For example, deep neural network ensembles predict relationships 
between chemical structure and reactivity \cite{Agrafiotis2002}, 
segment complex images with multiple objects \cite{Ju2017}, 
and are used in zero-shot as well as multiple choice 
learning \cite{Guzman2014,Ye2018}. 
Further, several winners and top performers on the ImageNet challenge 
are ensembles of neural networks \cite{Lee2015,Russakovsky2015}.
Ensembles function as collections of experts and have been shown, both 
theoretically and empirically, to improve generalization accuracy 
\cite{Drucker1993,Dietterich2000,Granitto2005,
Huggins2016,Ju2017,Lee2015,Russakovsky2015,Xu2014}.
For instance, by combining several image classification 
networks on the CIFAR-10, CIFAR-100, and SVHN data sets, ensembles 
can reduce the misclassification rate by up to $20$ percent, 
e.g., from $6$ percent to $4.5$ percent for ensembles of 
ResNets on CIFAR-10 \cite{Huang2017,Ju2017}. 


\begin{figure}[t]
	\includegraphics[width=\linewidth]{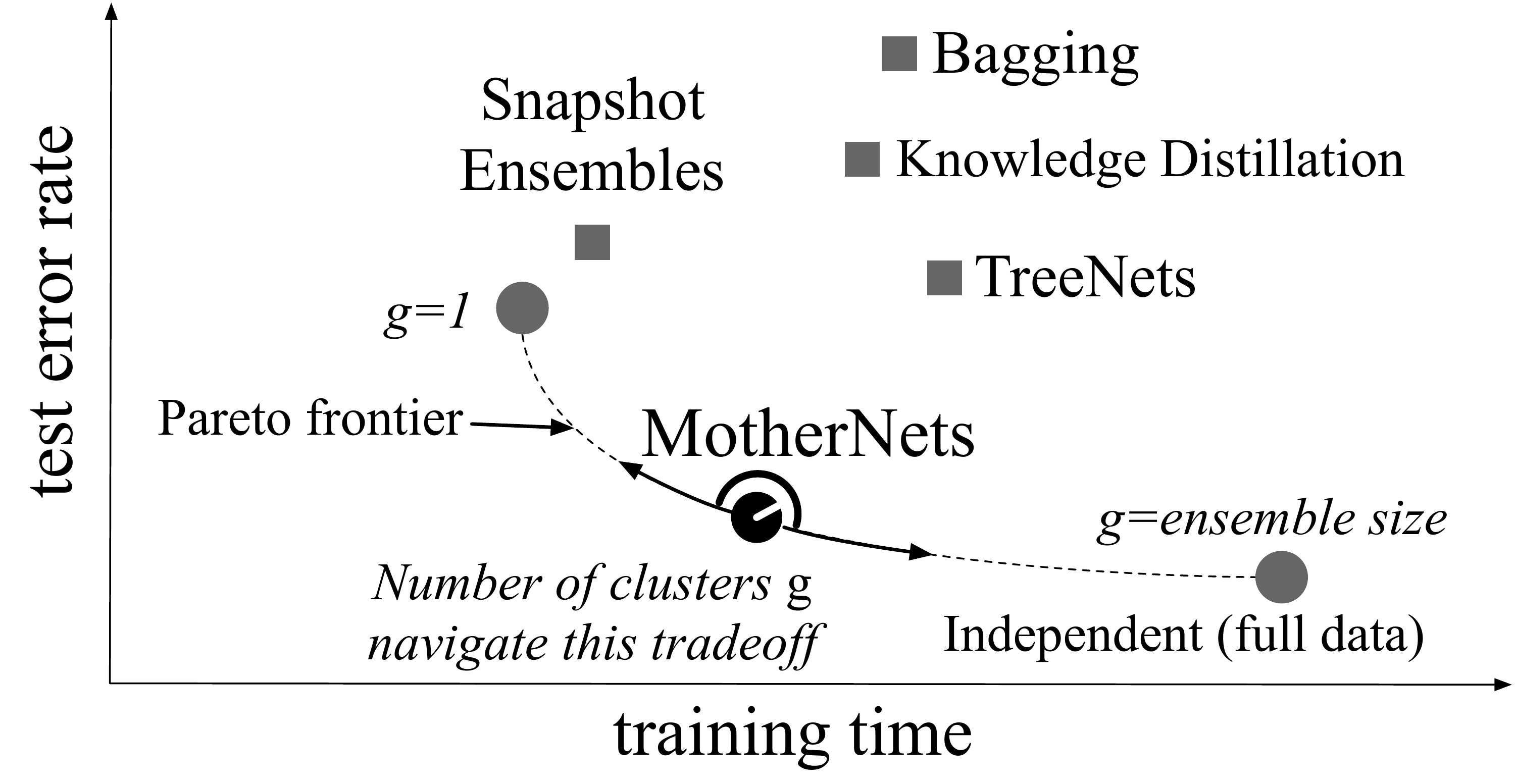}
	\caption{\label{fig:intro_conceptual}MotherNets establish a 
	new Pareto frontier for the accuracy-training time tradeoff as well 
	as navigate this tradeoff.}
	\vspace{-0.3in}
\end{figure}

\boldPara{The growing training cost.} 
Training ensembles of multiple deep neural networks takes 
a prohibitively large amount of time and computational 
resources. 
Even on high-performance hardware, a single deep neural network 
may take several days to train and this training cost grows linearly 
with the size of the ensemble as every neural network in the 
ensemble needs to be trained 
\cite{Szegedy2015,He2016,Huang2017a,Huang2017}. 
This problem persists even in the presence of multiple machines. 
This is because the holistic cost of training, in terms of buying 
or renting out these machines through a cloud service provider, 
still increases linearly with the ensemble size. 
The rising training cost is a bottleneck for numerous 
applications, especially when it is critical to quickly 
incorporate new data and to achieve a target accuracy. 
For instance, in one use case, where deep learning models 
are applied to detect Diabetic Retinopathy (a leading cause 
of blindness), newly labelled images become available every 
day. Thus, incorporating new data in the neural network 
models as quickly as possible is crucial in order to 
enable more accurate diagnosis for the immediately next 
patient \cite{Gulshan2016}.

\begin{table}[]
\vspace{-0.075in}
\caption{\label{tbl:comparison}
Existing approaches to train ensembles of deep neural 
networks are limited in speed, accuracy, diversity, 
and size.}
\vspace{0.1in}
\centering
\begin{tabular}{lcccc}
\hline
          & Fast & High & Diverse & Large\\ 
          & train. & acc. & arch.  & size\\ \hline 
Full data &      $\times$	&  \checkmark     & \checkmark   & $\times$\\
Bagging   &      $\sim$	 &  $\times$   & \checkmark     & $\times$\\
Knowledge Dist. & $\sim$      &  $\times$     & \checkmark & $\times$\\
TreeNets  &     $\sim$ &  $\sim$   & $\times$     & $\times$\\
Snapshot Ens.  &      \checkmark	&  \checkmark   & $\times$   & $\times$\\
\textbf{MotherNets}        &  {\checkmarkb}  &  \checkmarkb     & \checkmarkb & \checkmarkb\\
\hline 
\end{tabular}
\end{table}

\boldPara{Problem 1: Restrictive ensemble size.} 
Due to this prohibitive training cost, researchers and 
practitioners can only feasibly train and employ small 
ensembles \cite{Szegedy2015,He2016,Huang2017a,Huang2017}. 
In particular, neural network ensembles contain drastically fewer individual 
models when compared with ensembles of other machine learning 
methods. For instance, random decision forests, a popular 
ensemble algorithm, often has several hundreds of individual models 
(decision trees), whereas state-of-the-art ensembles of deep 
neural networks consist of around five networks 
\cite{He2016,Huang2017a,Huang2017,Oshiro2012,Szegedy2015}. 
This is restrictive since the generalization accuracy of an 
ensemble increases with the number of well-trained models it 
contains \cite{Oshiro2012,Bonab2016,Huggins2016}. 
Theoretically, for best accuracy, the size of the ensemble 
should be at least equal to the number of class 
labels, of which there could be thousands in modern 
applications \cite{Bonab2016}. 

\boldPara{Additional problems: Speed, accuracy, and diversity.} 
Typically, every deep neural network in an ensemble is initialized 
randomly and then trained individually using all training data 
(full data), or by using a random subset of the training data 
(i.e., bootstrap aggregation or bagging) \cite{Ju2017,Lee2015,Moghimi2016}.
This requires a significant amount of processing time and 
computing resources that grow linearly with the ensemble size.

To alleviate this linear training cost, two techniques have 
been recently introduced that generate a $k$ network ensemble 
from a single network: Snapshot Ensembles and TreeNets. 
Snapshot Ensembles train a single network and use its parameters 
at $k$ different points of the training process to instantiate 
$k$ networks that will form the target ensemble \cite{Huang2017}. 
Snapshot Ensembles vary the learning rate in a cyclical fashion, 
which enables the single network to converge to $k$ local 
minima along its optimization path.  
TreeNets also train a single network but this network 
is designed to branch out into $k$ sub-networks after the first 
few layers. Effectively every sub-network functions as a separate 
member of the target ensemble \cite{Stefan2015}. 

While these approaches do improve training time, they also come 
with two critical problems. First, the resulting ensembles are 
less accurate because they are less diverse compared to using 
$k$ different and individually trained networks.  Second, these 
approaches cannot be applied to state-of-the-art diverse 
ensembles. Such ensembles may contain arbitrary neural network 
architectures with structural differences to achieve increased 
accuracy (for instance, such as those used in the ImageNet 
competitions \cite{Lee2015,Russakovsky2015}). 

Knowledge Distillation provides a middle ground between separate 
training and ensemble generation approaches \cite{Hinton2015}. 
With Knowledge Distillation, an ensemble is trained by first 
training a large 
\textit{generalist} network and then distilling its knowledge to 
an ensemble of small \textit{specialist} networks that may have different 
architectures (by training them to mimic the probabilities produced 
by the larger network) \cite{Hinton2015,Li2017}. 
However, this approach results in limited improvement in training cost 
as distilling knowledge still takes around $70$ percent of the time 
needed to train from scratch. Even then, the ensemble networks are still 
closely tied to the same large network that they are distilled from. 
The result is significantly lower accuracy and diversity when compared 
to ensembles where every network is trained individually 
\cite{Hinton2015,Li2017}.

\begin{figure*}[t!]
	\includegraphics[width=\textwidth]{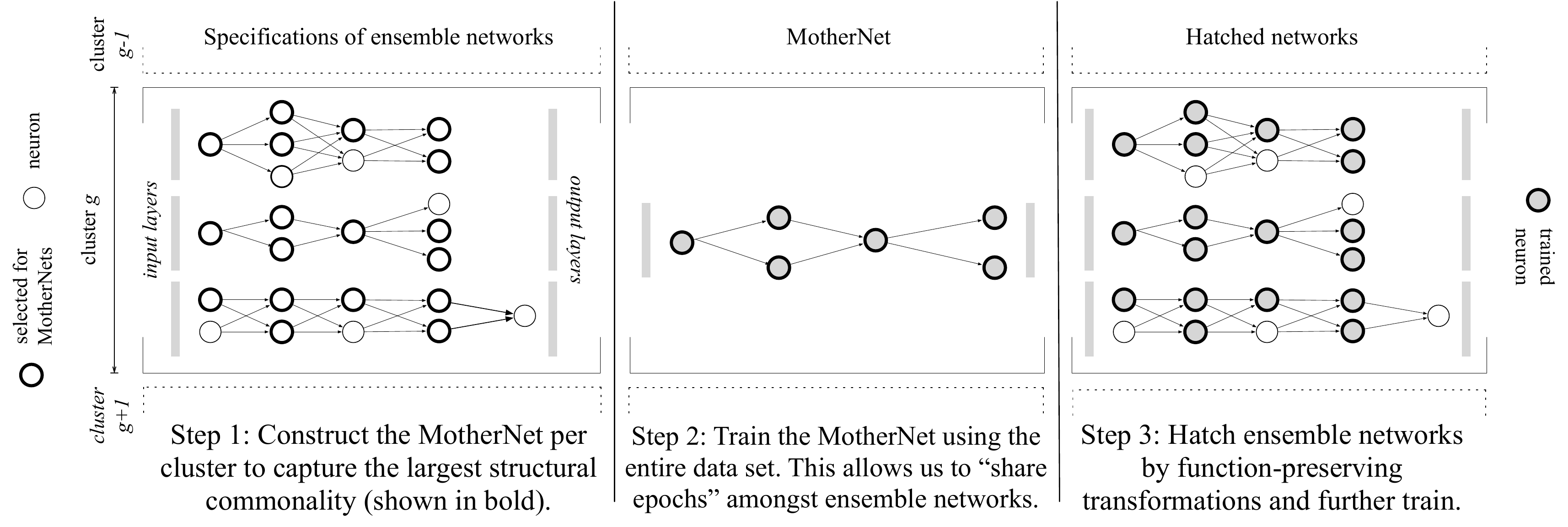}
	\vspace{-0.3in}
	\caption{\label{fig:example} MotherNets train an ensemble 
	of neural networks by first training a set of MotherNets and 
	transferring the function to the ensemble networks. The ensemble 
	networks are then further trained converging significantly 
	faster than training individually.}
	\vspace{-0.0in}
\end{figure*}

\boldPara{MotherNets.} 
We propose MotherNets, which enable rapid training of large 
feed-forward neural network ensembles. The core benefits 
of MotherNets are depicted in Table \ref{tbl:comparison}. 
MotherNets provide: 
(i) lower training time and better generalization accuracy than 
existing fast ensemble training approaches and 
(ii) the capacity to train large ensembles with diverse network 
architectures.

Figure \ref{fig:example} depicts the core intuition behind 
MotherNets: 
A MotherNet is a network that captures the maximum structural 
similarity between a cluster of networks (Figure 
\ref{fig:example} Step (1)). An ensemble may consist of one or more 
clusters; one MotherNet is constructed per cluster. 
Every MotherNet is trained to convergence using the full data 
set (Figure \ref{fig:example} Step (2)). 
Then, every target network in the ensemble is \textit{hatched} from 
its MotherNet using function-preserving transformations (Figure 
\ref{fig:example} Step (3)) ensuring that knowledge from the MotherNet 
is transferred to every network. The ensemble networks are then trained. 
They converge significantly faster compared to training from scratch 
(within tens of epochs).

The core technical intuition behind the MotherNets design 
is that it enables us to ``share epochs'' between the ensemble 
networks. At a lower level what this means is that the networks 
implicitly share part of the data movement and computation costs 
that manifest during training over the same data. This design 
draws intuition from systems techniques such as ``shared scans'' 
in data systems where many queries share data 
movement and computation for part of a scan over the same data 
\cite{Harizopoulos2005,Zukowski2007,Qiao2008,Arumugam2010,Candea2011,
Giannikis2012,Psaroudakis2013,Giannikis2014,Kester2017}.

\boldPara{Accuracy-training time tradeoff.} 
MotherNets do not train each network individually but ``source'' all networks 
from the same set of ``seed'' networks instead. This introduces some 
reduction in diversity and accuracy compared to an approach that 
trains all networks independently. 
There is no way around this. In practice, there is an 
intrinsic tradeoff between ensemble accuracy and training time. All 
existing approaches are affected by this and their design decisions 
effectively place them at a particular balance within this 
tradeoff \cite{Guzman2014,Lee2015,Huang2017}.

We show that MotherNets, strike a superior balance between accuracy 
and training time than all existing approaches. In fact, we show 
that MotherNets establish a new Pareto frontier for this tradeoff 
and that we can navigate this tradeoff. To achieve this, MotherNets 
cluster ensemble networks (taking into account both the topology 
and the architecture class) and train a separate MotherNet for 
each cluster. The number of clusters used (and thus the 
number of MotherNets) is a knob that helps navigate the 
training time vs. accuracy tradeoff. Figure  
\ref{fig:intro_conceptual} depicts visually the new tradeoff 
achieved by MotherNets.


\boldPara{Contributions.} We describe how to construct 
MotherNets in detail and how to trade accuracy for speed. 
Then through a detailed experimental evaluation with diverse 
data sets and architectures we demonstrate that MotherNets 
bring three benefits: 
(i) MotherNets establish a new Pareto frontier of the 
accuracy-training time tradeoff providing up to $2$ percent 
better absolute test error rate compared to fast ensemble 
training approaches at comparable or less training cost. 
(ii) MotherNets allow robust navigation of this new Pareto 
frontier of the tradeoff between accuracy and training time. 
(iii) MotherNets enable scaling of neural network ensembles to 
large sizes (100s of models) with practical training cost and 
increasing accuracy benefits.

We provide a web-based interactive demo as an additional 
resource to help in understanding the training process 
in MotherNets: \url{http://daslab.seas.harvard.edu/mothernets/}. 

\section{Rapid Ensemble Training}
\label{sec:main}

\boldPara{Definition: MotherNet.}
Given a cluster of $k$ neural networks $C = \{N_1, N_2, \dots N_{k}\}$,
where $N_i$ denotes the $i$-th neural network in $C$, the MotherNet $M_c$
is defined as the largest network from which all networks in $C$ can be
obtained through function-preserving transformations.
MotherNets divide an ensemble into one or more such network clusters 
and construct a separate MotherNet for each.

\textbf{Constructing a MotherNet for fully-connected networks.}
Assume a cluster $C$ of fully-connected neural networks. The input
and the output layers of $M_c$ have the same structure as all
networks in $C$, since they are all trained for the same task.
$M_c$ is initialized with as many hidden layers as the
shallowest network in $C$.
Then, we construct the hidden layers of $M_c$ one-by-one going from
the input to the output layer. The structure of the $i$-th
hidden layer of $M_c$ is the same as the $i$-th hidden layer 
of the network in $C$ with the least number of parameters at 
the $i$-th layer.
Figure \ref{fig:example} shows an example of how this process works
for a toy ensemble of two three-layered and one four-layered neural
networks.
Here, the MotherNet is constructed with three layers.
Every layer has the same structure as the layer with the
least number of parameters at that position (shown in
bold in Figure \ref{fig:example} Step (1)). In Appendix 
\ref{appendix:algorithms} we also include a pseudo-code 
description of this algorithm.

\begin{figure}[t]
		\centering
		\includegraphics[width=\linewidth]{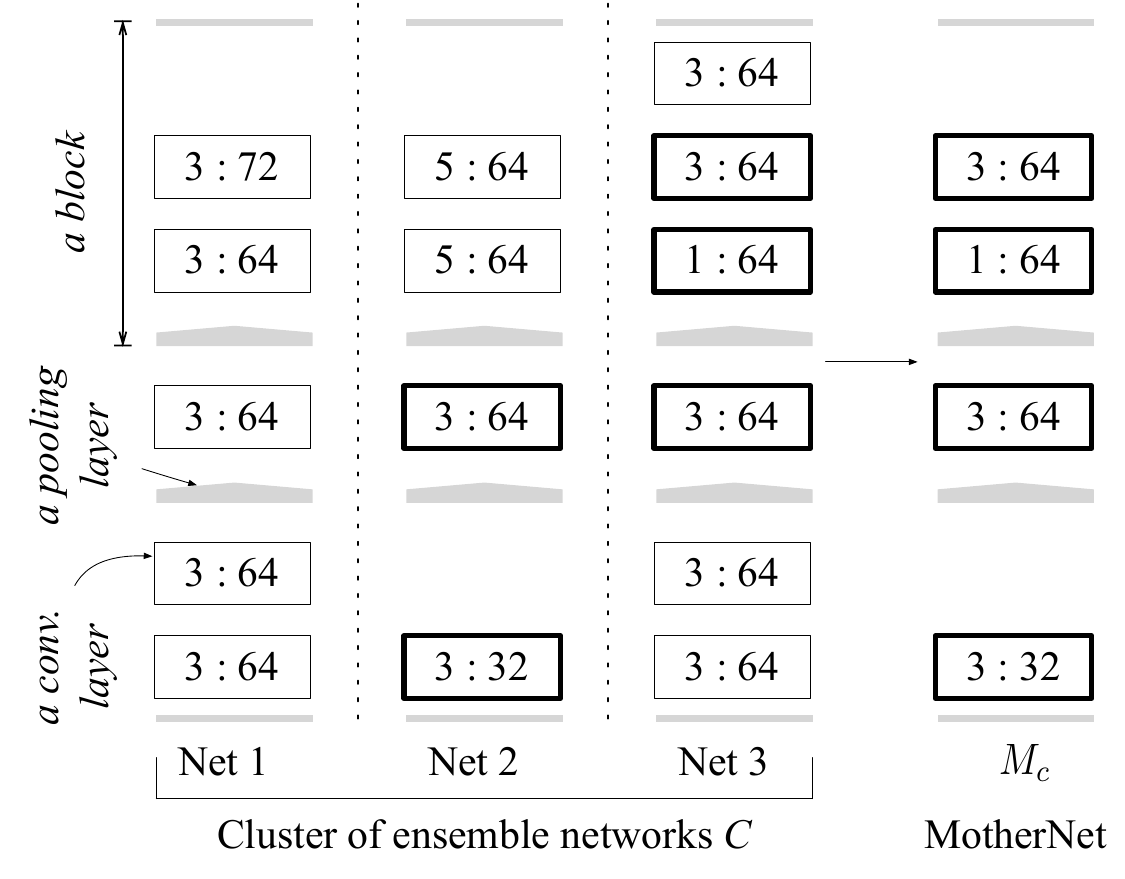}
		\vspace{-0.2in}
		\caption{\label{fig:example_block_by_block} Constructing
		MotherNet for convolutional neural networks block-by-block.
		For each layer, we select the layer with the least number
		of parameters from the ensemble networks (shown in bold
		rectangles) (Notation: \texttt{<filter\_width>:<filter\_number>}).}
		\vspace{-0.2in}
\end{figure}

\textbf{Constructing a MotherNet for convolutional networks.}
Convolutional neural network architectures consist of blocks of
one or more convolutional layers separated by pooling layers
\cite{He2016,Shazeer2017,Simonyan2014,Szegedy2015}.
These blocks are then followed by another block of one or more
fully-connected layers.
For instance, VGGNets are composed of five blocks of convolutional
layers separated by max-pooling layers, whereas, DenseNets consist
of four blocks of densely connected convolutional layers.
For convolutional networks, we construct the MotherNet $M_c$
block-by-block instead of layer-by-layer.
The intuition is that deeper or wider variants of such networks
are created by adding or expanding layers within individual blocks
instead of adding them all at the end of the network. For instance,
VGG-C (with 16 convolutional layers) is obtained by adding one layer
to each of the last three blocks of VGG-B (with 13 convolutional
layers) \cite{Simonyan2014}.
To construct the MotherNet for every block, we select as many
convolutional layers to include in the MotherNet as the network
in $C$ with the least number of layers in that block. Every layer
within a block is constructed such that it has the least number
of filters and the smallest filter size of any layer at the same
position within that block. An example of this process is shown 
in Figure \ref{fig:example_block_by_block}. Here, we construct 
a MotherNet for three convolutional neural networks block-by-block. 
For instance, in the first block, we include one convolutional 
layer in the MotherNet having the smallest filter width and 
the least number of filters (i.e., 3 and 32 respectively).
In Appendix \ref{appendix:algorithms} we also include a 
pseudo-code description of this algorithm.

\boldPara{Constructing MotherNets for ensembles of neural networks
with different sizes and topologies.}
By construction, the overall size and topology (sequence of
layer sizes) of a MotherNet is limited by the smallest network
in its cluster. If we were to assign a single cluster to all
networks in an ensemble that has a large difference in size and
topology between the smallest and the largest networks, there will
be a correspondingly large difference between at least one ensemble
network and the MotherNet. This may lead to a scenario where the
MotherNet only captures an insignificant amount of commonality.
This would negatively affect performance as we would not be 
able to share significant computation and data movement costs 
across the ensemble networks. This property is directly correlated 
with the size of the MotherNet.

In order to maintain the ability to share costs in diverse 
ensembles, we partition such an ensemble into $g$ clusters, 
and for every cluster, we construct and train a separate 
MotherNet. 
To perform this clustering, the $m$ networks in the ensemble
$E = \{N_1, N_2, \dots N_{m}\}$ are represented as vectors
$E_v = \{V_1, V_2, \dots V_{m}\}$ such that $V_i^j$ stores the size
of the $j$-th layer in $N_i$. These vectors are
zero-padded to a length of $\max(\{|N_1|, |N_2|, \dots |N_{m}|\})$
(where $|N_i|$ is the number of layers in $N_{i}$).
For convolutional neural networks, these vectors are
created by first creating similarly zero-padded sub-vectors
per block and then concatenating the sub-vectors to get the
final vector. In this case, to fully represent convolutional
layers, $V_i^j$ stores a 2-tuple of filter sizes and number of
filters.

Given a set of vectors $E_v$, we create $g$ clusters 
using the balanced K-means algorithm while minimizing the 
Levenshtein distance between the vector representation of 
networks in a cluster and its MotherNet 
\cite{Levenshtein1966,Macqueen1967}. The Levenshtein or 
the edit distance between two vectors is the minimum number 
of edits -- insertions, deletions, or substitutions -- needed 
to transform one vector to another. 
By minimizing this distance, we ensure that, for every cluster, the 
ensemble networks can be obtained from their cluster's 
MotherNet with the minimal amount of edits constrained on 
$g$.
During every iteration of the K-means algorithm, instead of 
computing centers of candidate clusters, we construct MotherNets 
corresponding to every cluster. Then, we use the edit distance 
between these MotherNets and all networks to perform cluster 
reassignments.

\boldPara{Constructing MotherNets for ensembles of diverse
architecture classes.}
An individual MotherNet is built for a cluster of networks that
belong to a single architecture class. Each architecture class
has the property of function-preserving navigation. This is to 
say that given any member of this class, we can build another 
member of this class with more parameters but having the same function.
Multiple types of neural networks fall under the same architecture
class \cite{Cai2017}. For instance, we can build a single MotherNet
for ensembles of AlexNets, VGGNets, and Inception Nets as well as 
one for DenseNets and ResNets.
To handle scenarios when an ensemble contains members from diverse
architecture classes i.e., we cannot navigate the entire set of
ensemble networks in a function-preserving manner, we build a
separate MotherNet for each class (or a set of MotherNets if each
class also consists of networks of diverse sizes).

Overall, the techniques described in the previous paragraphs 
allow us to create $g$ MotherNets for an ensemble, being able 
to capture the structural similarity across diverse networks 
both in terms of architecture and topology. We now describe 
how to train an ensemble using one or more MotherNets to help 
share the data movement and computation costs amongst the 
target ensemble networks.

\boldPara{Training Step 1: Training the MotherNets.}
First, the MotherNet for every cluster is trained from scratch
using the entire data set until convergence. This allows the
MotherNet to learn a good core representation of the data.
The MotherNet has fewer parameters than any of the networks
in its cluster (by construction) and thus it takes less time per epoch
to train than any of the cluster networks.

\boldPara{Training Step 2: Hatching ensemble networks.} Once the
MotherNet corresponding to a cluster is trained, the next step is
to generate every cluster network through a sequence of
function-preserving transformations that allow us to expand the
size of any feed-forward neural network, while ensuring that the
function (or mapping) it learned is preserved \cite{Chen2015}.
We call this process \textit{hatching} and there are two distinct
approaches to achieve this:
Net2Net increases the capacity of the given network by adding identity
layers or by replicating existing weights \cite{Chen2015}.
Network Morphism, on the other hand, derives sufficient and necessary
conditions that when satisfied will extend the network while preserving
its function and provides algorithms to solve for those conditions
\cite{Wei2016,Wei2017}.

In MotherNets, we adopt the first approach i.e., Net2Net. Not only 
is it conceptually simpler but in our experiments we observe that 
it serves as a better starting point for further training of the 
expanded network as compared to Network Morphism. 
Overall, function-preserving transformations are readily applicable to 
a wide range of feed-forward neural networks including VGGNets, ResNets, FractalNets, DenseNets, and Wide ResNets 
\cite{Chen2015,Wei2016,Wei2017,Huang2017a}. As such MotherNets is 
applicable to all of these different network architectures. In 
addition, designing function-preserving transformations is an 
active area of research and better transformation techniques 
may be incorporated in MotherNets as they become available. 

Hatching is a computationally inexpensive process that takes negligible
time compared to an epoch of training \cite{Wei2016}. This is because
generating every network in a cluster through function preserving
transformations requires at most a single pass on layers in its
MotherNet.
%

\boldPara{Training Step 3: Training hatched networks.}
To explicitly add diversity to the hatched networks, we randomly
perturb their parameters with gaussian noise before further
training. This breaks symmetry after hatching and it is a standard
technique to create diversity when training ensemble networks
\cite{Hinton2015,Stefan2015,Wei2016,Wei2017}. Further, adding 
noise forces the hatched networks to be in a different 
part of the hypothesis space from their MotherNets.

The hatched ensemble networks are further trained converging
significantly faster compared to training from scratch. This fast
convergence is due to the fact that by initializing every ensemble
network through its MotherNet, we placed it in a good position in the
parameter space and we need to explore only for a relatively small
region instead of the whole parameter space. We show that hatched
networks typically converge in a very small number epochs.

{We experimented with both full data and bagging to train
hatched networks. We use full data because given the small 
number of epochs needed for the hatched networks, bagging 
does not offer any significant advantage in speed while 
it hurts accuracy.

%
\renewcommand{\arraystretch}{1.35}
\begin{table*}[ht]
\centering
\caption{{We experiment with ensembles of various sizes and neural
network architectures.}\label{tbl:ens}}
\vspace{0.1in}
\begin{tabular}{p{15mm}p{90mm}p{10mm}p{25mm}p{10mm}}

	\hline
	\textbf{Ensemble}&
	\textbf{Member networks}&
	\textbf{Param.}&
	\textbf{SE alternative}&
	\textbf{Param.}\\
	\hline
	\textbf{V5}&
	VGG $13$, $16$, $16A$, $16B$, and $19$ from the 
	VGGNet paper \cite{Simonyan2014}
	& 682M & VGG-16 $\times$ 5 & 690M \\
	\textbf{D5} &
	Two variants of DenseNet-40
	(with $12$ and $24$ convolutional filters per layer) and
	three variants of DenseNet-100 (with $12$, $16$, and $24$
	filters per layer) \cite{Huang2017a}
	& 17M & DenseNet-60 $\times$ 5 & 17.3M \\
	\textbf{R10}&
	Two variants each of ResNet 20, 32, 44, 56, and 110 from the 
	ResNet paper \cite{He2016}
	& 327M & R-56 $\times$ 10 & 350M \\
	\textbf{V25}&
	$25$ variants of VGG-16 with distinct architectures 
	created by progressively varying one layer from VGG16
	in one of three ways: (i) increasing the number of filters,
	(ii) increasing the filter size, or (iii) applying both (i)
	and (ii) 
	& 3410M & VGG-16 $\times$ 25 & 3450M \\
	\textbf{V100}&
	$100$ variants of VGG-$16$ created as described above 
	& 13640M & VGG-16 $\times$ 100 & 13800M \\
	\hline
\end{tabular}
\end{table*}

\boldPara{Accuracy-training time tradeoff.}
{
MotherNets can navigate the tradeoff between accuracy and training
time by controlling the number of clusters $g$, which in turn
controls how many MotherNets we have to train independently
from scratch.
For instance, on one extreme if $g$ is set to $m$, then every network
in $E$ will be trained independently, yielding high accuracy at the
cost of higher training time.
%
On the other extreme, if $g$ is set to one then, all ensemble networks
have a shared ancestor and this process may yield networks that are
not as diverse or accurate, however, the training time will be low.}

MotherNets expose $g$ as a tuning knob. As we show in our 
experimental analysis, MotherNets achieve a new Pareto 
frontier for the accuracy-training cost tradeoff which is 
a well-defined convex space. That is, with every step in 
increasing $g$ (and consequently the number of independently trained 
MotherNets) accuracy does get better at the cost of some 
additional training time and vice versa. Conceptually this 
is shown in Figure \ref{fig:intro_conceptual}. This convex 
space allows robust and predictable navigation of the tradeoff. 
For example, unless one needs best accuracy or best training time 
(in which case the choice is simply the extreme values of g), 
they can start with a single MotherNet and keep adding MotherNets 
in small steps until the desired accuracy is achieved or the 
training time budget is exhausted. This process can further be 
fine-tuned using known approaches for hyperparameter tuning
methods such as bayesian optimization, training on sampled data, 
or learning trajectory sampling \cite{Goodfellow2016}.

\boldPara{Parallel training.} MotherNets create a new schedule 
for ``sharing epochs'' amongst networks of an ensemble but the 
actual process of training in every epoch remains unchanged. 
As such, state-of-the-art approaches for distributed training 
such as parameter-server \cite{Dean2012} and asynchronous gradient 
descent \cite{Gupta2016,Iandola2016} can be applied to fully utilize 
as many machines as available during any stage of MotherNets' 
training.

\boldPara{Fast inference.} 
MotherNets can also be used to improve inference time by 
keeping the MotherNet parameters shared across the hatched 
networks. We describe this idea in Appendix \ref{appendix:shared}.

\begin{figure*}[]
	\begin{subfigure}{0.218\textwidth}
		\includegraphics[trim=0 0 5 0,clip=true, width=\textwidth]
			{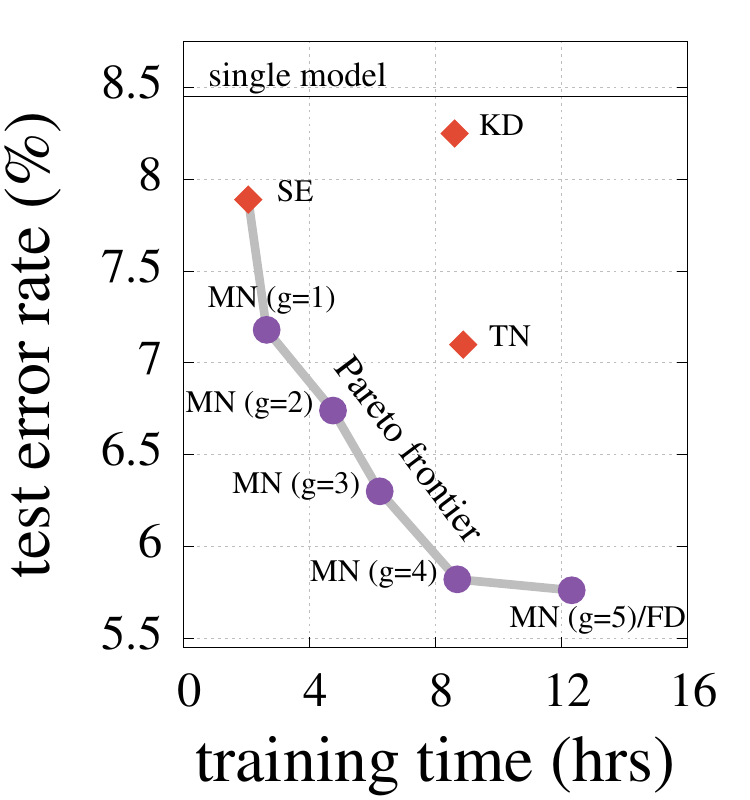}
		\caption{V5 (C-10)}
		\label{exp:V5_all}
	\end{subfigure}
	\begin{subfigure}{0.1915\textwidth}
		\includegraphics[trim=25 0 5 0,clip,width=\textwidth]
			{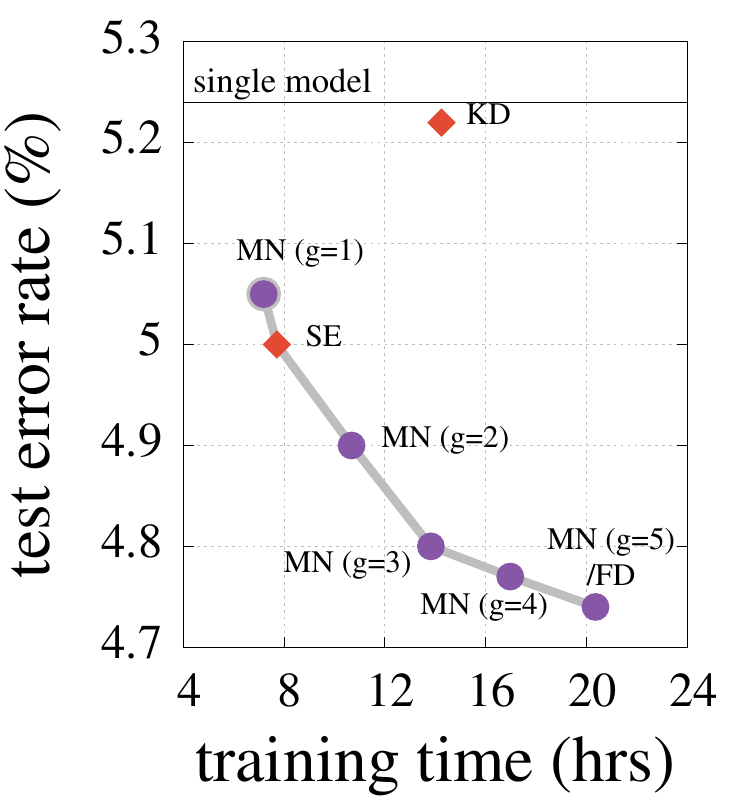}
		\caption{D5 (C-10)}
		\label{exp:Dense5_all}
	\end{subfigure}
	\begin{subfigure}{0.1915\textwidth}
		\includegraphics[trim=25 0 5 0,clip, width=\textwidth]
			{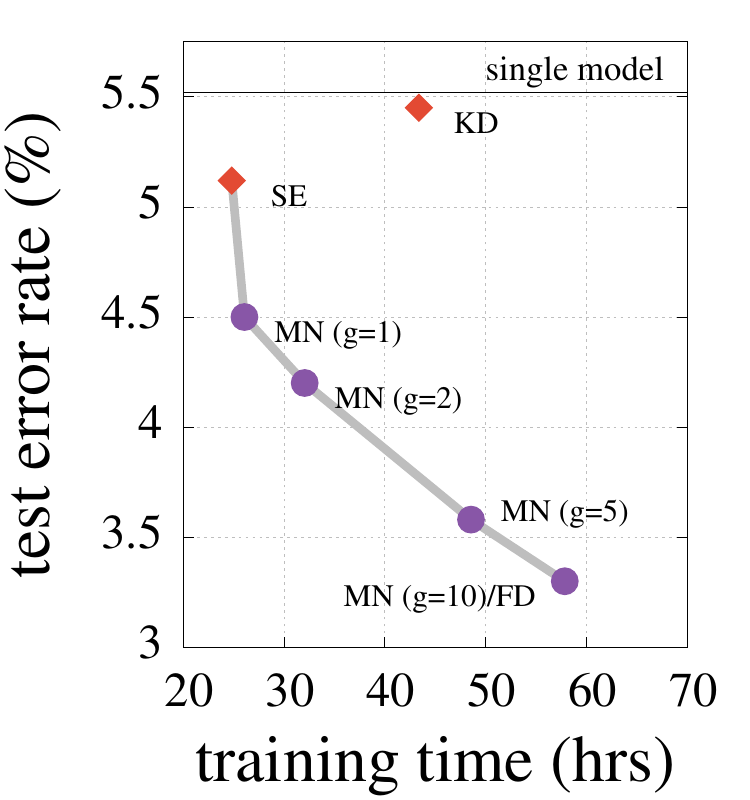}
		\caption{R10 (C-10)}
		\label{exp:R_25_CIFAR_10_tradeoff}
	\end{subfigure}
	\begin{subfigure}{0.1915\textwidth}
		\includegraphics[trim=25 0 5 0,clip, width=\textwidth]
			{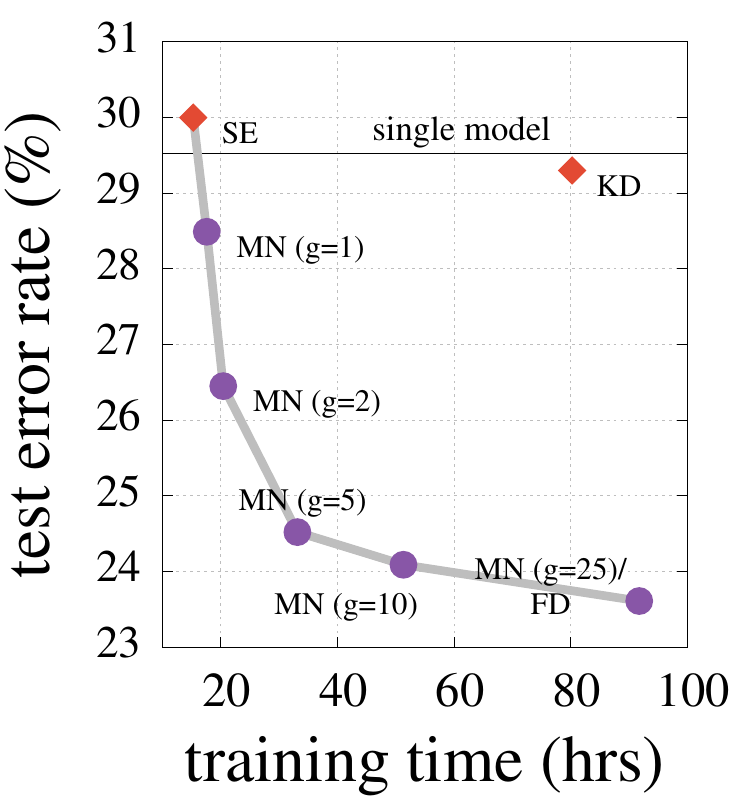}
		\caption{V25 (C-100)}
		\label{exp:V_25_CIFAR_100_tradeoff}
	\end{subfigure}
\begin{subfigure}{0.1915\textwidth}
		\includegraphics[trim=25 0 5 0,clip, width=\textwidth]
			{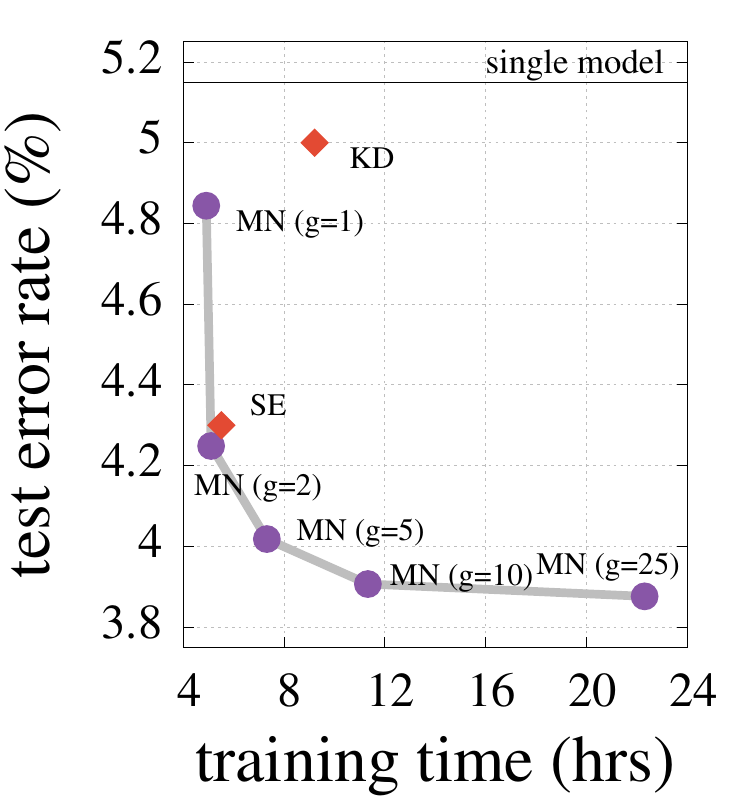}
		\caption{V25 (SVHN)}
		\label{exp:V_25_SVHN_tradeoff}
\end{subfigure}
\caption{\centering MotherNets provide consistently better accuracy-training 
time tradeoff when compared with existing fast ensemble training 
approaches across various data sets, architectures, and ensemble 
sizes.}
\vspace{-0.1in}
\label{exp:all_tradeoff}
\end{figure*}

\section{Experimental Analysis}
\label{sec:experiments}

We demonstrate that MotherNets enable a better training 
time-accuracy tradeoff than existing fast ensemble training 
approaches across multiple data sets and architectures. We 
also show that MotherNets make it more realistic to utilize 
large neural network ensembles. 

\boldPara{Baselines.}
We compare against five state-of-the-art methods spanning both 
techniques that train all ensemble networks individually, i.e., 
Full Data (FD) and Bagging (BA), as well as approaches that 
generate ensembles by training a single network, i.e., 
Knowledge Distillation (KD), Snapshot Ensembles (SE), and 
TreeNets (TN).

\boldPara{Evaluation metrics.} We capture both the training 
cost and the resulting accuracy of an ensemble.
For the training cost, we report the wall clock time as well 
as the monetary cost for training on the cloud.
For ensemble test accuracy, we report the test error 
rate under the widely used ensemble-averaging method 
\cite{Van2007,Guzman2012,Guzman2014,Stefan2015}.
Experiments with alternative inference methods (e.g., super 
learner and voting \cite{Ju2017}) showed that the method we 
use does not affect the overall results in terms of comparing 
the training algorithms.

\boldPara{Ensemble networks.}
We experiment with ensembles of various convolutional architectures
such as VGGNets, ResNets, Wide ResNets\footnote{For experiments 
with Wide ResNets, see Appendix \ref{appendix:FGE}.}, and DenseNets.
{Ensembles of these architectures have been extensively used
to evaluate fast ensemble training approaches \cite{Lee2015,Huang2017}.}
Each of these ensembles are composed of networks having diverse
architectures as described in Table \ref{tbl:ens}.

To provide a fair comparison with SE (where the snapshots have to
be from the same network architecture), we create snapshots having
comparable number of parameters to each of the ensembles described
above. This comparable alternatives we used for SE are also
summarized in Table \ref{tbl:ens}.

For TN, we varied the number of shared layers and found that
sharing the $3$ initial layers provides the best accuracy. This
is similar to the optimal proportion of shared layers in the
TreeNets paper \cite{Lee2015}. 
TN is not applicable to DenseNets or ResNets as it is designed 
only for networks without skip-connections \cite{Lee2015}. 
We omit comparison with TN for such ensembles.

\boldPara{Training setup.}
For all training approaches we use stochastic 
gradient descent with a mini-batch size of
$256$ and batch-normalization. All weights
are initialized by sampling from a standard normal distribution. 
Training data is randomly shuffled before every training epoch.
The learning rate is set to $0.1$ with the
exception of DenseNets. For DenseNets, we use a learning rate of 
$0.1$ to train MotherNets and $0.01$ to train hatched networks. 
This is inline with the learning rate decay used in the DenseNets 
paper \cite{Huang2017a}. For FD, KD, TN, and MotherNets, we stop 
training if the training accuracy does not improve for 15 epochs. 
For SE we use the optimized training setup proposed in the original 
paper \cite{Huang2017}, starting with an initial learning rate of 
0.2 and then training every snapshot for 60 epochs.

\boldPara{Data sets.} We experiment with a diverse array of 
data sets: SVHN, CIFAR-10, and CIFAR-100 
\cite{Krizhevsky2009,Netzer2011}.
The SVHN data set is composed of images of house numbers and
has ten class labels. There are a total of $99$K images. We
use $73$K for training and $26$K for testing.
The CIFAR-10 and CIFAR-100 data sets have $10$ and $100$ class
labels respectively corresponding to various images of everyday
objects. There are a total of $60$K images -- $50$K training and
$10$K test images.

\boldPara{Hardware platform.} All experiments are run on the same 
server with Nvidia Tesla V100 GPU.

\subsection{Better training time-accuracy tradeoff}

We first show how MotherNets strike an overall superior
accuracy-training time tradeoff when compared to existing
fast ensemble training approaches.

Figure \ref{exp:all_tradeoff} shows results across all our 
test data sets and ensemble networks. All graphs in Figure \ref{exp:all_tradeoff} 
depict the tradeoff between training time needed versus accuracy 
achieved. The core observation from Figure \ref{exp:all_tradeoff} 
is that across all datasets and networks, MotherNets help establish 
a new Pareto frontier of this tradeoff. The different versions of 
MotherNets shown in Figure \ref{exp:all_tradeoff} represent different 
numbers of clusters used ($g$). When $g$=1, we use a single MotherNet, 
optimizing for training time, while when $g$ becomes equal to the 
ensemble size, we optimize for accuracy 
(effectively this is equal to FD as every network is trained 
independently in its own cluster).

The horizontal line at the top of each graph indicates the 
accuracy of the best-performing single model in the ensemble 
trained from scratch. This serves as a benchmark and, in the 
vast majority of cases, all approaches do improve 
over a single model even when they have to sacrifice on accuracy 
to improve training time. MotherNets is consistently 
and significantly better than that benchmark.

Next we discuss each individual training approach and how it 
compares to MotherNets.

\boldPara{MotherNets vs. KD, TN, and BA.}
MotherNets (with $g$=1) is $2\times$ to $4.2\times$ faster 
than KD and results in up to $2$ percent better test accuracy.
KD suffers in terms of accuracy because its ensemble networks
are more closely tied to the base network 
as they are trained from the output of the same network. 
KD's higher training cost is because distilling is 
expensive. Every network starts from scratch and is trained 
on the data set using a combination of empirical loss and 
the loss from the output of the teacher network. We 
observe that distilling a network still takes around $60$ 
to $70$ percent of the time required to train it using just 
the empirical loss.

To achieve comparable accuracy to MotherNets (with $g$=1),
TN requires up to $3.8\times$ more training time on $V5$.
In the same time budget, MotherNets can train with $g$=4
providing over one percent reduction in test error rate.
The higher training time of TN is due to the fact that 
it combines several networks together to create a monolithic 
architecture with various branches. We observe that training 
this takes a significant time per epoch as well as requires 
more epochs to converge. Moreover, TN does not generalize 
to neural networks with skip-connections.
%

Figure \ref{exp:all_tradeoff} does not show results for BA 
because it is an outlier. BA takes on average 73 percent of 
the time FD needs to train but results in significantly higher 
test error rate than any of the baseline approaches including 
the single model. Compared to BA, MotherNets is on average 3.6 
$\times$ faster and results in significantly better accuracy -- 
up to 5.5 percent lower absolute test error rate. These 
observations are consistent with past studies that show how 
BA is ineffective when training deep neural networks as it reduces 
the number of unique data items seen by individual networks 
\cite{Lee2015}. 

Overall, the low test error rate of MotherNets when compared to KD, TN, and
BA stems from the fact that transferring the learned function
from MotherNets to target ensemble networks provides a good
starting point as well as introduces regularization for further
training. This also allows hatched ensemble networks to converge
significantly faster, resulting in overall lower training time.

\boldPara{Training time breakdown.}
To better understand where the time goes during the training 
process, Figure \ref{exp:Dense5_CIFAR10_training_time} provides 
the time breakdown per ensemble network. We show this for the 
D5 ensemble and compare MotherNets (with $g$=1) with individual 
training approaches FD, BA, and KD. While other approaches 
spend significant time training each network, MotherNets, can 
train these networks very quickly after having trained the 
core MotherNet (black part in the MotherNets stacked bar in 
Figure \ref{exp:Dense5_CIFAR10_training_time}). We observe 
similar time breakdown across all ensembles in our experiments. 

\begin{figure}[t]
	\begin{minipage}[t]{0.2375\textwidth}
		\includegraphics[width=\textwidth]
			{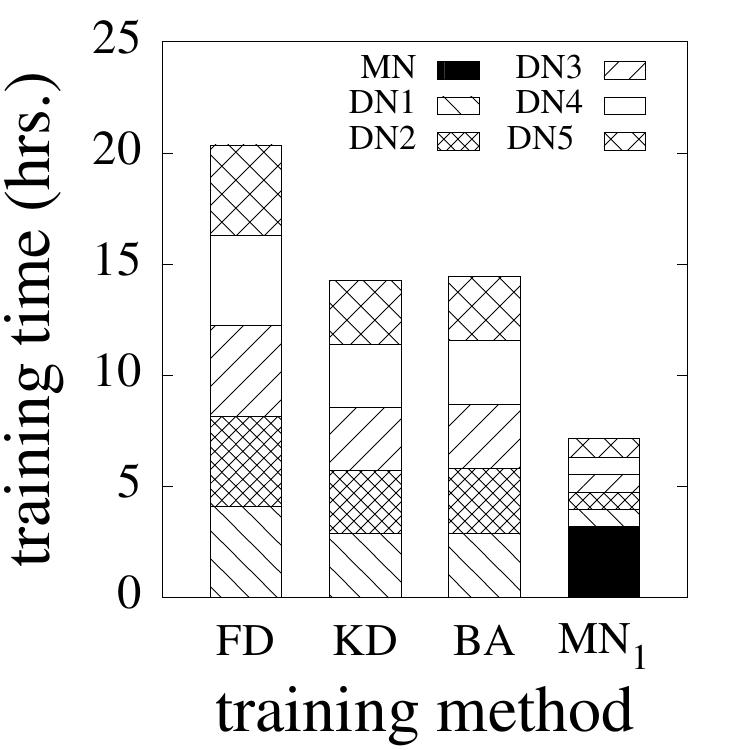}
		\begin{minipage}{0.9 \textwidth}
			\caption{{MotherNets train ensemble networks 
			significantly faster after having trained the 
			MotherNet (shown in black).}}
			\label{exp:Dense5_CIFAR10_training_time}
		\end{minipage}
	\end{minipage}
	\begin{minipage}[t]{0.2375\textwidth}
		\includegraphics[width=\textwidth]
			{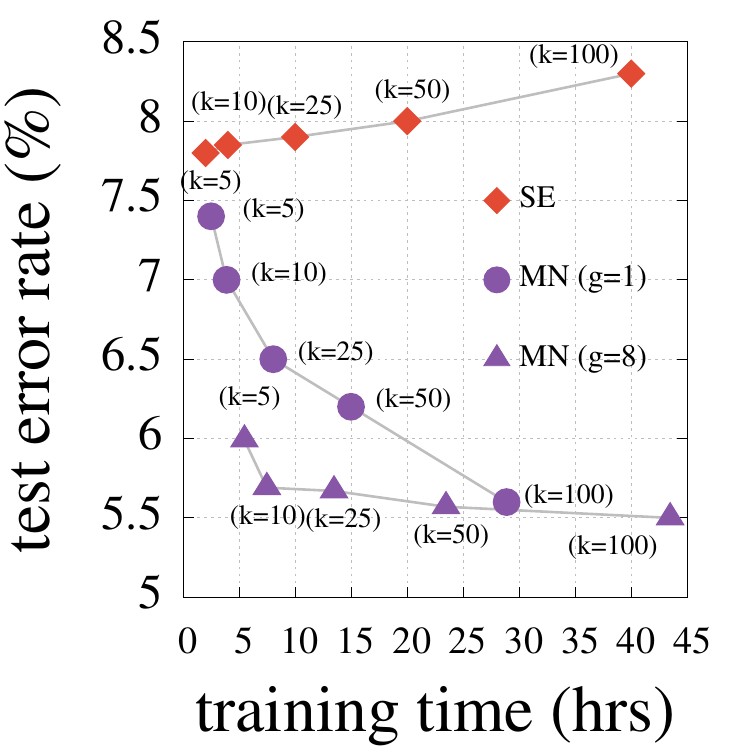}
		\begin{minipage}{0.9 \textwidth}
			\caption{{As the size of the ensemble grows, 
			MotherNets scale better than SE both in terms 
			of training time and accuracy achieved.}}
			\label{exp:V_100_CIFAR_10_tradeoff}
		\end{minipage}
	\end{minipage}
	\vspace{-0.2in}
\end{figure}


\subsection{MotherNets vs. SE and scaling to large ensembles}

Across all experiments in Figure \ref{exp:all_tradeoff}, SE is the 
closest baseline to MotherNets. In effect, SE is part of the very 
same Pareto frontier defined by MotherNets in the accuracy-training 
cost tradeoff. That is, it represents one more valid point that can 
be useful depending on the desired balance. 
For example, in Figure \ref{exp:V5_all} (for V5 CIFAR-10), SE 
sacrifices nearly one percent in test error rate compared to 
MotherNets (with $g$=1) for a small improvement in training cost. We 
observe similar trends in Figure \ref{exp:R_25_CIFAR_10_tradeoff} 
and \ref{exp:V_25_CIFAR_100_tradeoff}). 
In Figure \ref{exp:Dense5_all}, SE achieves a balance that is in 
between MotherNets with one and two clusters. However, when training 
V25 on SVHN (Figure \ref{exp:V_25_SVHN_tradeoff}) SE is in fact 
outside the Pareto frontier as it is both slower and achieves 
worst accuracy. 

Overall, MotherNets enables drastic improvements in either accuracy 
or training time compared to SE by being able to control and navigate 
the tradeoff between the two.

\begin{table}[t]
\caption{\label{tbl:oracle}MotherNets (with $g$=1) give better oracle
test accuracy compared to Snapshot ensembles.}
\vspace{0.1in}
\centering
\begin{tabular}{llllll}
\hline
            & \textbf{\begin{tabular}[c]{@{}l@{}}V5 \\ C10\end{tabular}} & \textbf{\begin{tabular}[c]{@{}l@{}}D5\\ C10\end{tabular}} & \textbf{\begin{tabular}[c]{@{}l@{}}R10\\ C10\end{tabular}} & \textbf{\begin{tabular}[c]{@{}l@{}}V25 \\ C100\end{tabular}} & \textbf{\begin{tabular}[c]{@{}l@{}}V25 \\ SVHN\end{tabular}} \\ \hline
\textbf{MN} & 96.71                                                      & 97.43                                                     & 98.61                                                      & 87.5                                                         & 97.17                                                        \\
\textbf{SE} & 96.03                                                      & 96.91                                                     & 97.11                                                      & 86.9                                                         & 97.3                                                         \\ \hline
\end{tabular}
\vspace{-0.2in}
\end{table}

\textbf{Oracle accuracy.} 
Also, Table \ref{tbl:oracle} shows that MotherNets (with $g$=1) 
enable better oracle 
test accuracy when compared with SE across all our experiments. 
This is the accuracy if an \textit{oracle} were to pick the 
prediction of the most accurate network in the ensemble per 
test element \cite{Guzman2012,Guzman2014,Stefan2015}. Oracle 
accuracy is an upper bound for the accuracy 
that any ensemble inference technique could achieve. 
This metric is also used to evaluate the utility of 
ensembles when they are applied to solve Multiple Choice 
Learning (MCL) problems \cite{Guzman2014,Lee2016,Brodie2018}.

\boldPara{Scaling to very large ensembles.}
As we discussed before, large ensembles help improve accuracy 
and thus ideally we would like to scale 
neural network ensembles to large number of models as it happens 
for other ensembles such as random forests 
\cite{Oshiro2012,Bonab2016,Bonab2017}. Our previous results 
were for small to medium ensembles of 5, 10 or 25 networks. 
We now show that when it comes to larger ensembles, MotherNets 
dominate SE in both how accuracy and training 
time scale. 

Figure \ref{exp:V_100_CIFAR_10_tradeoff} shows results as we 
increase the number of networks up to a hundred variants of 
VGGNets trained on CIFAR-10. 
For every point in Figure \ref{exp:V_100_CIFAR_10_tradeoff}, 
$k$ indicates the number of networks. For MotherNets we plot 
results for the time-optimized version with $g$=1, as well as 
with $g$=8.

Figure \ref{exp:V_100_CIFAR_10_tradeoff} 
shows that as the size of the ensemble grows, MotherNets 
scale much better in terms of training time. Toward the end (for 100 
networks), MotherNets train more than 10 hours faster (out of 
40 total hours needed for SE). The training time of MotherNets 
grows at a much smaller rate because once the MotherNet has been 
trained, it takes $40$ percent less time to train a hatched network 
than the time it takes to train one snapshot. 

In addition, Figure \ref{exp:V_100_CIFAR_10_tradeoff} shows that 
MotherNets does not only scale better in terms of training time, 
but also it scales better in 
terms of accuracy. As we add more networks to the ensemble, 
MotherNets keeps improving its error rate by nearly 2 percent 
while SE actually becomes worse by more than 0.5 percent. The 
declining accuracy of SE as the size of the 
ensemble increases has also been observed in the past, where by 
increasing the number of snapshots above six results in degradation 
in performance \cite{Huang2017}.

Finaly, Figure \ref{exp:V_100_CIFAR_10_tradeoff} shows that 
different cluster settings for MotherNets allow us to achieve 
different performance balances while still providing robust 
and predictable navigation of the tradeoff. In this case, 
with $g$=8 accuracy improves consistently across all points 
(compared to $g$=1) at the cost of extra training time.

\begin{wrapfigure}{r}{0.25\textwidth}
		\vspace{-0.25in}
		\includegraphics[trim=9 5 0 0, clip, width=0.25\textwidth]
			{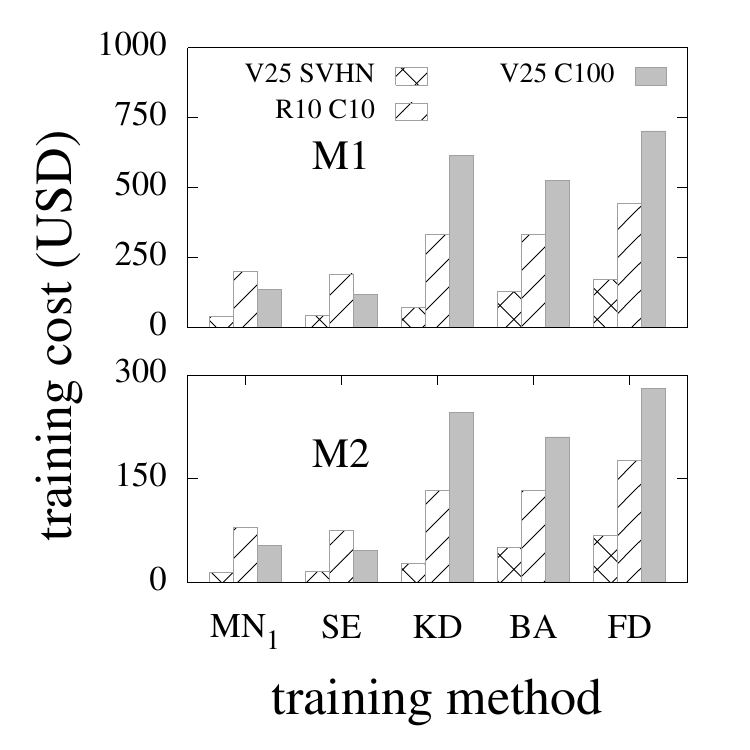}
			\vspace{-0.15in}
		\caption{{Training cost (USD)}}
		\label{exp:training_cost}
		\vspace{-0.25in}
\end{wrapfigure}
\subsection{Improving cloud training cost}
One approach to speed up training of large ensembles is to 
utilize more than one machines. For example, we could train 
$k$ individual networks in parallel using $k$ machines. 
While this does save time, the holistic cost in terms of 
energy and resources spent is still linear to the ensemble 
size. 

One proxy for capturing the holistic cost is to look at the 
amount of money one has to pay on the cloud for training a 
given ensemble. In our next experiment, we compare all 
approaches using this proxy. 
Figure \ref{exp:training_cost} shows the cost (in USD)
of training on four cloud instances across two cloud service
providers:
(i) M1 that maps to AWS P2.xlarge and Azure NC6, and
(ii) M2 that maps to AWS P3.2xlarge and Azure NCv3. M1 is priced
at USD $0.9$ per hour and M2 is priced at USD $3.06$ per hour
for both cloud service providers
\cite{awsPricing,azurePricing}.

Training time-optimized MotherNets
provide significant reduction in training cost (up to $3 \times$) as
it can train a very large ensemble in a fraction of the training time 
compared to other approaches.

\subsection{Diversity of model predictions}

Next, we analyze how diversity of ensembles produced by
MotherNets compares with SE and FD.

\boldPara{Ensembles and predictive diversity.} Theoretical results 
suggest that ensembles of models perform better when the
models' predictions on a single example are less correlated. This
is true under two assumptions: (i) models have equal correct
classification probability and (ii) the ensemble uses majority vote
for classification \cite{Krogh1994, Rosen1996, Kuncheva2003}.
Under ensemble averaging, no analytical proof that negative correlation
reduces error rate exists, but lower correlation between models can
be used to create a smaller upper bound on incorrect classification
probability. More precise statements and their proofs are given in
Appendix \ref{appendix:covariance}.


\boldPara{Rapid ensemble training methods.} For MotherNets, as well as for
all other compared techniques for ensemble training, the training procedure binds the models together to decrease training time. This can have two negative effects compared to independent training of models:
\begin{enumerate}
\item by changing the model's architecture or training pattern, the technique affects each model's prediction quality (the model's marginal prediction accuracy suffers)
 \item by sharing layers (TN), attempted softmax values (KD), or training epochs (SE, MN), the training technique creates positive correlations between model errors.
 \end{enumerate}
We compare here the magnitude of these two effects forMotherNets and Snapshot Ensembles when compared to independent training of each model on CIFAR-10 using V5. 

\begin{figure}
\captionsetup{width=\textwidth}
\includegraphics[width=0.485\textwidth] {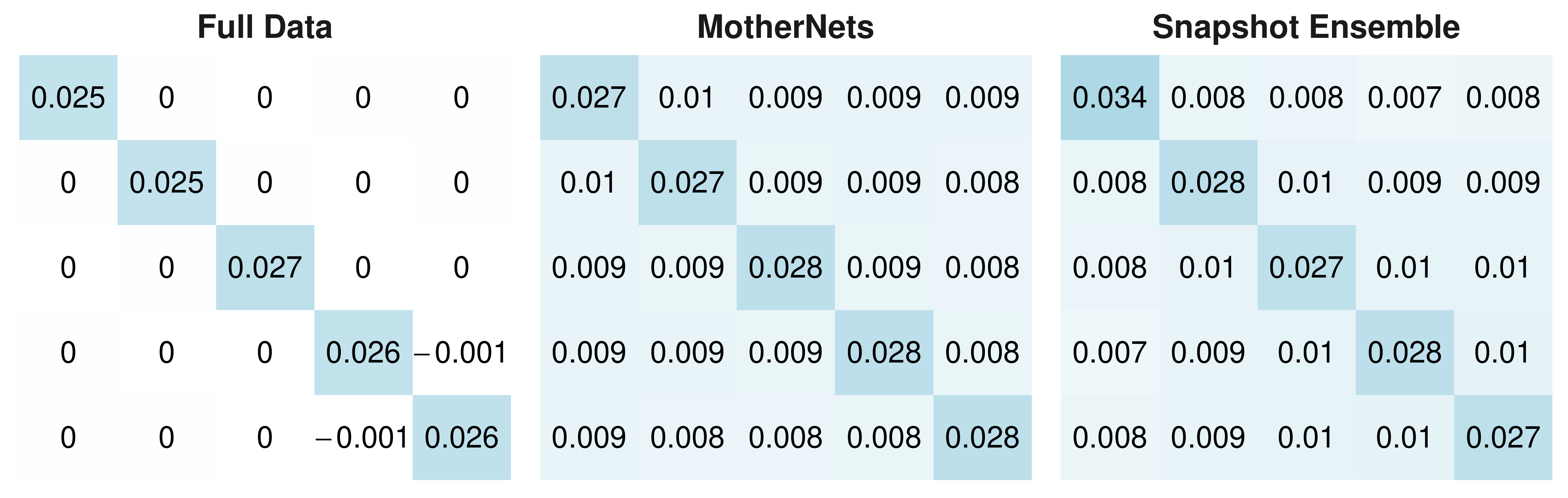}
\caption{{MotherNets (with $g$=1) train ensembles with lower model 
covariances compared to Snapshot Ensembles.}}\label{fig:covar}
\end{figure}

\boldPara{Individual model quality. } For both SE and MN, the individual model accuracy drops, but the effect is more pronounced in SE than MN. The mean misclassification percentage of the individual models for V5 using FD, MN and SE are $8.1\%$, $8.4\%$ and $9.8\%$ respectively. The poor performance of SE in this area is due to its difficulty in consistently hitting performant local minima, either because it overfits to the training data when trained for a long time or because its early snapshots need to be far away from the final optimum to encourage diversity.

\boldPara{Model variance.} Our goal in assessing variance is to 
see how the training procedure affects how models in the ensemble 
correlate with each other on each example. To do this, we train each 
of the five models in $V5$ five times under MN, SE, and FD. Letting 
$Y_{ij}$ be the softmax of the correct model on test example $j$ using model $i$, we then estimate $Var(Y_{ij})$ for each $i,j$ and $Cov(Y_{ij}, Y_{i'j})$ for each $i, i', j$ with $i \neq i'$ using the sample variance and covariance. To get a single number for a model, instead of one for each test example, we then average across all test examples, i.e. $Cov(Y_{i}, Y_{i'}) = \frac{1}{n} \sum_{j=1}^{n} Cov(Y_{ij}, Y_{i'j})$. For total variance numbers for the ensemble, we perform the same procedure on $Y_{j} = \frac{1}{5} \sum_{i=1}^{5} Y_{ij}$.

Figure \ref{fig:covar} shows the results. As expected, independent 
training between the models in FD makes their corresponding covariance 
0 and provides the greatest overall variance reduction for the ensemble, 
with ensemble variance at $0.0051$. For both SE and MN, the covariance of 
separate models is non-zero at around $0.009$ per pair of models; however, 
it is also significantly less than the variance of a single model. As a 
result, both MN and SE provide significant variance reduction compared to 
a single model. Whereas a single model has variance around $0.026$, MN 
and SE provide ensemble variance of $0.0125$ and $0.0130$ respectively.

\boldPara{Takeaways.} Since both SE and MN train nearly as fast as a 
single model, they provide variance reduction in prediction at very 
little training cost.  Additionally, for MN, at the cost of higher 
training time, one can create more clusters and thus make the training 
of certain models independent of each other, zeroing out many of the 
covariance terms and reducing the overall ensemble variance. When compared 
to each other, MN with $g$=1 and SE have similar variance numbers, 
with MN slightly lower, but MotherNets has a substantial increase in 
individual model accuracy when compared to Snapshot Ensembles. 
As a result, its overall ensemble performs better.

\boldPara{Additional results.} 
We demonstrate in Appendix \ref{appendix:shared} how MotherNets can 
improve inference time by $2 \times$. 
In Appendix \ref{appendix:parallel}, we show how the relative behavior 
of MotherNets remains the same when training using multiple GPUs. 
Finally, in Appendix \ref{appendix:FGE} we provide experiments 
with Wide ResNets and demonstrate how MotherNets provide better 
accuracy-training time tradeoff when compared with Fast Geometric 
Ensembles.

\section{Related Work}

In this section, we briefly survey additional (but orthogonal) 
ensemble training techniques beyond Snapshot Ensembles, 
TreeNets, and Knowledge Distillation.

\boldPara{Parameter sharing.} 
Various related techniques share parameters between different 
networks during ensemble training and, in doing so, improve 
training cost. 
One interpretation of techniques such as Dropout and Swapout 
is that, during training, they create several networks with 
shared weights within a single network. Then, they implicitly 
ensemble them during inference 
\cite{Wan2013,Srivastava2014,Huang2016,Singh2016,Huang2017}. 
Our approach, on the other hand, captures the structural 
similarity in an ensemble, where members have different and 
explicitly defined neural network architectures and trains it. 
Overall, this enables us to effectively 
combine well-known architectures together within an ensemble. 
Furthermore, implicit ensemble techniques (e.g., dropout and 
swapout) can be used as training optimizations to improve 
upon the accuracy of individual networks trained in MotherNets 
\cite{Srivastava2014,Singh2016}.


\boldPara{Efficient deep network training.} 
Various algorithmic techniques target fundamental bottlenecks 
in the training process \cite{Niu2011,Brown2016,Bottou2016}. 
Others apply system-oriented techniques to reduce memory 
overhead and data movement \cite{de2017,Jain2018}. 
Recently, specialized hardware is being developed to improve 
performance, parallelism, and energy consumption of neural 
network training \cite{prabhakar2016,de2017,jouppi2017}.
All techniques to improve upon training efficiency of individual 
neural networks are orthogonal to MotherNets and in fact directly 
compatible. This is 
because MotherNets does not make any changes to the core 
computational components of the training process. 
In our experiments, we do utilize some of the widely applied 
training optimizations such as batch-normalization and 
early-stopping. The advantage that MotherNets bring on top 
of these approaches is that we can further reduce the total 
number of epochs that are required to train an ensemble. 
This is because a set of MotherNets will train for the 
structural similarity present in the ensemble. 

\section{Conclusion}

We present MotherNets which enable training of large and diverse 
neural network ensembles while being able to navigate a new Pareto 
frontier with respect to accuracy and training cost. The core 
intuition behind MotherNets is to reduce the number of epochs 
needed to train an ensemble by capturing the structural 
similarity present in the ensemble and training for it once. 

\section{Acknowledgments}
We thank reviewers for their valuable feedback. We also thank 
Chang Xu for building the web-based demo and all DASlab members 
for their help. This work was partially funded by Tableau, Cisco, 
and the Harvard Data Science Institute.  




\bibliographystyle{./style/sysml2019}
\bibliography{./common/bibliography/systems,./common/bibliography/ref,./common/bibliography/reference,./common/bibliography/library,./common/bibliography/to-be-added}


\renewcommand\thesection{\Alph{section}}
\setcounter{section}{0}
\renewcommand\thesubsection{\thesection.\arabic{subsection}}
\setcounter{subsection}{0}
\renewcommand\thefigure{\Alph{figure}}
\setcounter{figure}{0}
\renewcommand\thetable{\alph{table}}
\setcounter{table}{0}
\renewcommand\thealgorithm{\Alph{algorithm}}
\setcounter{algorithm}{0}

\newpage
\section*{Appendix}

\newcommand\mycommfont[1]{\small\ttfamily\textcolor{blue}{#1}}
\SetCommentSty{mycommfont}
\SetNoFillComment
\SetKwProg{Fn}{Function}{}{}

\begin{algorithm}[]
	\caption{Constructing the MotherNet for 
	fully-connected neural networks}\label{alg:construct_mn}
	\BlankLine
	\textbf{Input:} E: ensemble networks in one cluster;\\
	\textbf{Initialize:} M: empty MotherNet;  
	\BlankLine
	
	\acmt{set input/output layer sizes}
	M.input.num\_param $\gets$ E[0].input.num\_param;\\ 
	M.output.num\_param $\gets$ E[0].output.num\_param;\\
	M.num\_layers $\gets$ getShallowestNetwork(E).num\_layers;
	
	\BlankLine	
	\acmt{set hidden layer sizes}
	
	\For{$i \gets$ 0 \ldots M.num\_layers-1}{
		
		M.layers[i].num\_param $\gets$ getMin(E,i);
	}
	return M;\\
	
	\acmt{Get the min. size layer at posn}
	
	\Fn{getMin(E,posn)}{
		min $\gets$ E[0].layers[posn].num\_param;\\
		\For{$j \gets$ 0 \ldots len(E)-1}{
			\If{E[j].layers[posn].num\_param $<$ min}{
				min $\gets$ E[j].layers[posn].num\_param
			}
		}
		return min;
	}	
\end{algorithm}

\section{Algorithms for constructing MotherNets}
\label{appendix:algorithms}
We outline algorithms for constructing the MotherNet 
given a cluster of neural networks. We describe the algorithms 
for both fully-connected and convolutional neural networks. 

\boldPara{Fully-Connected Neural Networks.} 
{
Algorithm \ref{alg:construct_mn} describes how to construct the 
MotherNet for a cluster of fully-connected neural networks. 
We proceed layer-by-layer selecting the layer with the least 
number of parameters at every position.}

\boldPara{Convolutional Neural Networks.} 
{Algorithm \ref{alg:construct_mn_conv} provides a detailed strategy 
to construct the MotherNet for a cluster of convolutional neural 
networks. We proceed block-by-block, where each block is composed 
of multiple convolutional layers. 
The MotherNet has as many blocks as the network with the least 
number of blocks. Then, for every block, we proceed layer-by-layer 
and construct the MotherNet layer at every position as follows: 
First, we compute the least number of convolutional filters and 
convolutional filter sizes at that position across all ensemble 
networks. Let these be $F_{min}$ and $S_{min}$ respectively. 
Then, in MotherNet, we include a convolutional layer with 
$F_{min}$ filters of $S_{min}$ size at that position.

\begin{algorithm*}[t!]
	\caption{Constructing the MotherNet for convolutional neural networks 
	block-by-block.}
	\label{alg:construct_mn_conv}
	\BlankLine
	\textbf{Input:} E: ensemble of convolutional networks in one cluster;\\
	\textbf{Initialize:} M: empty MotherNet;  
	\BlankLine
	\acmt{set input/output layer sizes and number of blocks}
	
	M.input.num\_param $\gets$ E[0].input.num\_param; \\
	M.output.num\_param $\gets$ E[0].output.num\_param; \\
	M.num\_blocks $\gets$ getShallowestNetwork(E).num\_blocks;	
	
	\BlankLine
	\acmt{set hidden layers block-by-block}
	
	\For{$k \gets$ 0 \ldots M.num\_blocks-1}{
		M.block[k].num\_hidden $\gets$ getShallowestBlockAt(E,k).num\_hidden;
		\acmt{select the shallowest block}
		
		\For{$i \gets$ 0 \ldots M.block[k].num\_hidden-1}{
		
			M.block[k].hidden[i]..num\_filters, M.block[k].hidden[i]..filter\_size $\gets$ getMin(E,k,i)
		}
	}
	return M;\\
	
	\acmt{Get minimum number of filters and filter size at posn}
	
	\Fn{getMin(E,blk,posn)}{
		\BlankLine
		min\_num\_filters $\gets$ E[0].block[blk].hidden[posn].num\_filters;\\
		min\_filter\_size $\gets$ E[0].block[blk].hidden[posn].filter\_size;\\
		\For{$j \gets$ 0 \ldots len(E)}{
			\If{E[j].block[blk].hidden[posn].num\_filters $<$ min\_num\_filters}{
				min\_num\_filters $\gets$ 
				E[j].block[blk].hidden[posn].num\_filters;
			}
			\If{E[j].block[blk].hidden[posn].filter\_size $<$ min\_filter\_size}{
				min\_filter\_size $\gets$ 
				E[j].block[blk].hidden[posn].filter\_size;
			}
		}
		return min\_num\_filters, min\_filter\_size;
	}	
\end{algorithm*}


\label{appendix:algorithms}

\section{Model Covariance and Ensemble Predictive Accuracy}

We can analyze how model covariance effects ensemble performance by using Chebyshev's Inequality to bound the chance that a model predicts an example incorrectly. By showing that lower covariance between models makes this bound on the probability smaller, we give an intuitive reason why ensembles with lower covariance between models perform better. The proof shows as well that the average model's predictive accuracy is important; finally, no assumptions need to be made for the proof to hold. The individual models can be of different quality and have different chances of getting each example correct.

Given a fixed training dataset, let $Y_{i}$ be the softmax value of model $i$ in the ensemble for the correct class, and let $\hat Y = \frac{1}{m} \sum_{i=1}^{m} Y_{i}$ be the ensemble's average softmax value on the correct class. Both are random variables with the randomness of $\hat Y$ and $Y_{i}$ coming through the randomness of neural network training. Under the mild assumption that $E[\hat Y] > \frac{1}{2}$, so that the a one vs. all softmax classifier would say on average that the correct class is more likely, than Chebyshev's Inequality bounds the probability of incorrect prediction. Namely, the correct prediction is made with certainty if $\hat Y \geq \frac{1}{2}$ and so the probability of incorrect prediction is less than 
\vspace{-0.2in}
$$P(|\hat Y - E[\hat Y]| \geq E[\hat Y] - \frac{1}{2}) \leq \frac{Var(\hat Y)}{(E[\hat Y] - \frac{1}{2})^{2}}$$

From the form of the equation, we immediately see that keeping the average model accuracy $E[Y_{i}]$ high is important, and that degradation in model quality can offset reductions in variance. Since the variance of $\hat Y$ decomposes into $\frac{1}{m^{2}}(\sum_{i=1}^{m} Var(Y_{i}) + \sum_{i \neq i'} Cov(Y_{i}, Y_{i'}))$, we see that low model covariance keeps the variance of the ensemble low, and that models which have which have high covariance with other models provides little benefit to the ensemble.

\label{appendix:covariance}



We explain how MotherNets improve the efficiency of 
ensemble inference.

\boldPara{Ensemble inference.} 
Inference in an ensemble of neural networks proceeds as 
follows: 
First, the data item (e.g., an image or a feature vector) is 
passed through every network in the ensemble. These forward 
passes produce multiple predictions -- one prediction 
for every network in the ensemble. The prediction of the 
ensemble is then computed by combining the individual 
predictions using some averaging or voting function. 
As the size of the ensemble grows, the inference cost in terms of 
memory and time required for inference increases linearly. This 
is because for every additional ensemble network, we need to maintain 
its parameters as well as do an additional forward pass on them. 


\begin{figure}[ht!]
		\includegraphics[width=\columnwidth]{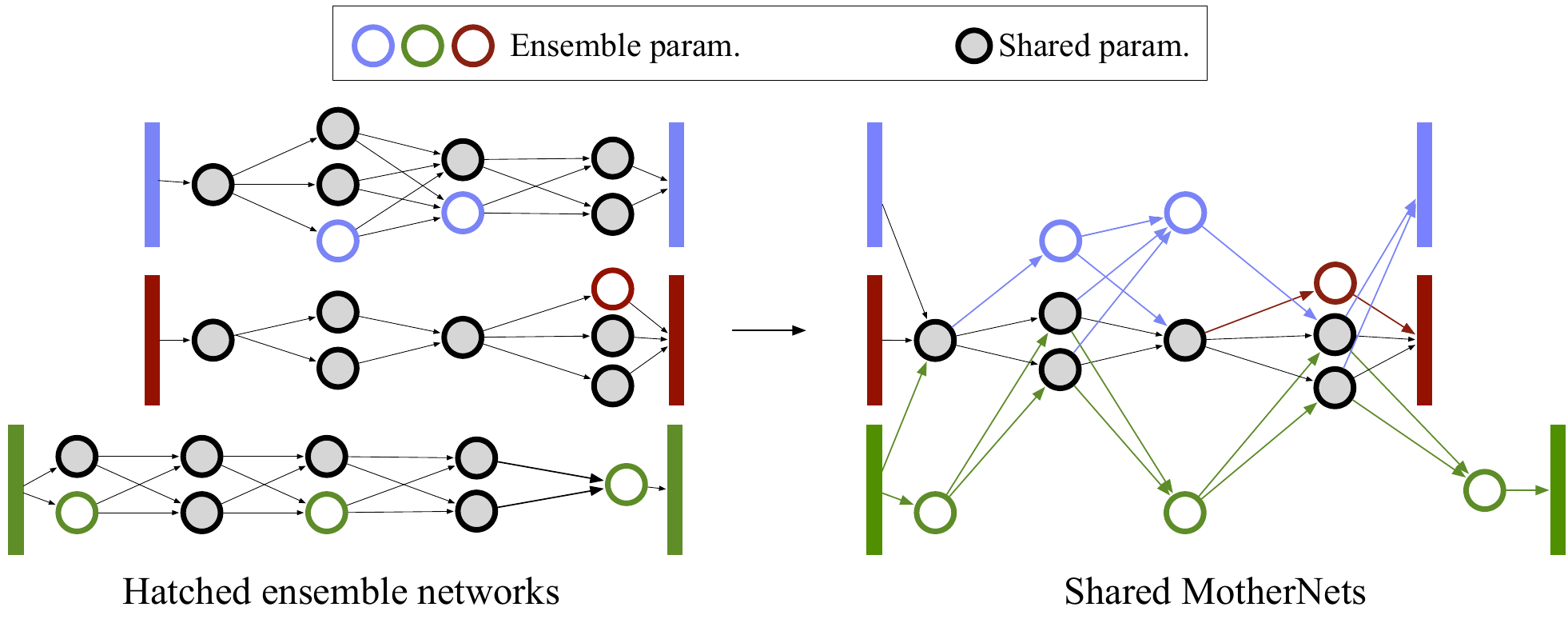}
		\caption{To construct a shared-MotherNet, parameters originating 
		from the MotherNet are combined together in the ensemble.}
		\label{fig:inference}
\end{figure}
\section{Shared-MotherNets}
\boldPara{Shared-MotherNets.} 
We introduce shared-MotherNets to reduce inference time and 
memory requirement of ensembles trained through MotherNets. 
In shared-MotherNets, after the process of hatching (step 2 
from \S 2), the parameters originating from the MotherNet 
are incrementally trained in a shared manner.  
This yields a neural network ensemble with a single copy of 
MotherNet parameters reducing both inference time and 
memory requirement. 
%

\begin{figure*}[ht!]
\captionsetup{width=0.34\textwidth}
\centering
	\begin{minipage}[t]{0.245\textwidth}
	\includegraphics[trim = 12 0 0 0, width=\textwidth]
			{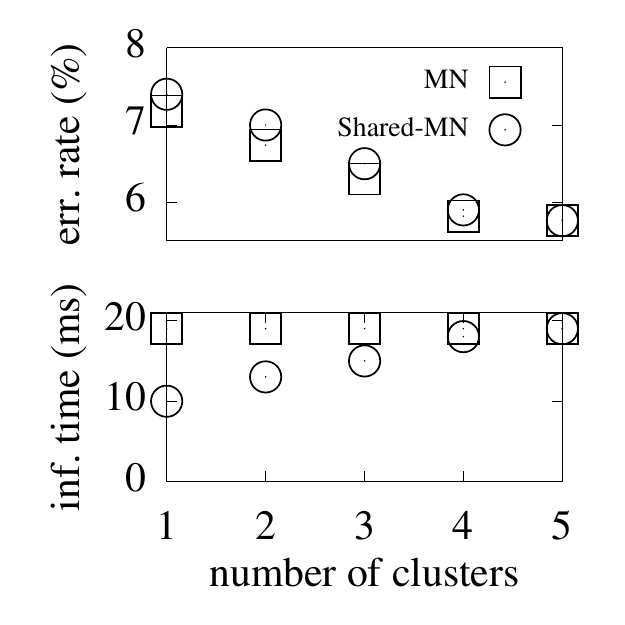}
	\hspace{0.1in}
	\begin{minipage}{0.9\textwidth}
		\caption{
		\label{exp:shared_mn}
		Shared MotherNets improve inference time by 
		2$\times$ for the V5 ensemble.}
	\end{minipage}
	\end{minipage}
	\begin{minipage}[t]{0.245\textwidth}
	\includegraphics[width=\textwidth]
			{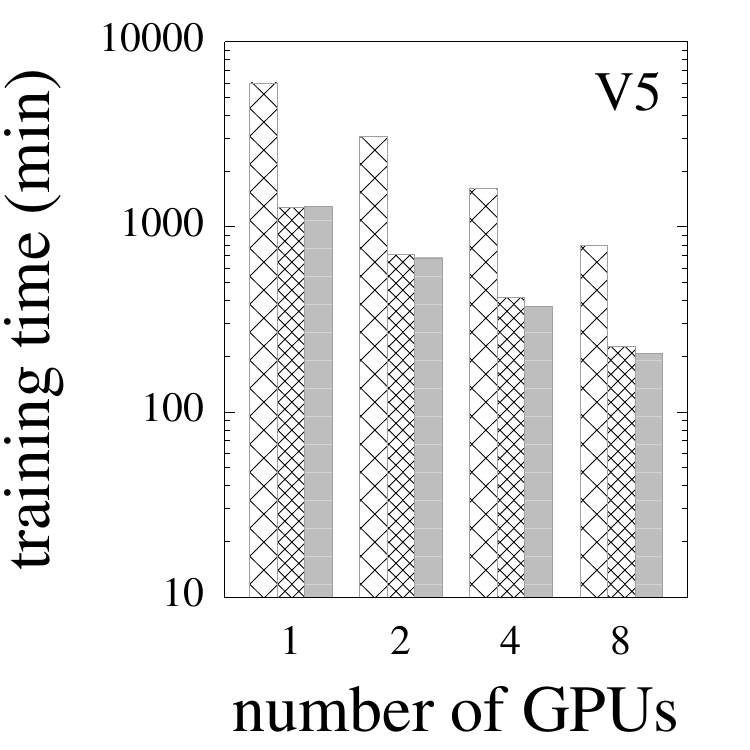}
			\begin{minipage}{0.9\textwidth}
				\caption{\label{exp:parallel_V5}MotherNets 
				continue to improve training cost in 
				parallel settings (V5).}
			\end{minipage}
	\end{minipage}
	\begin{minipage}[t]{0.245\textwidth}
	\includegraphics[width=\textwidth]
			{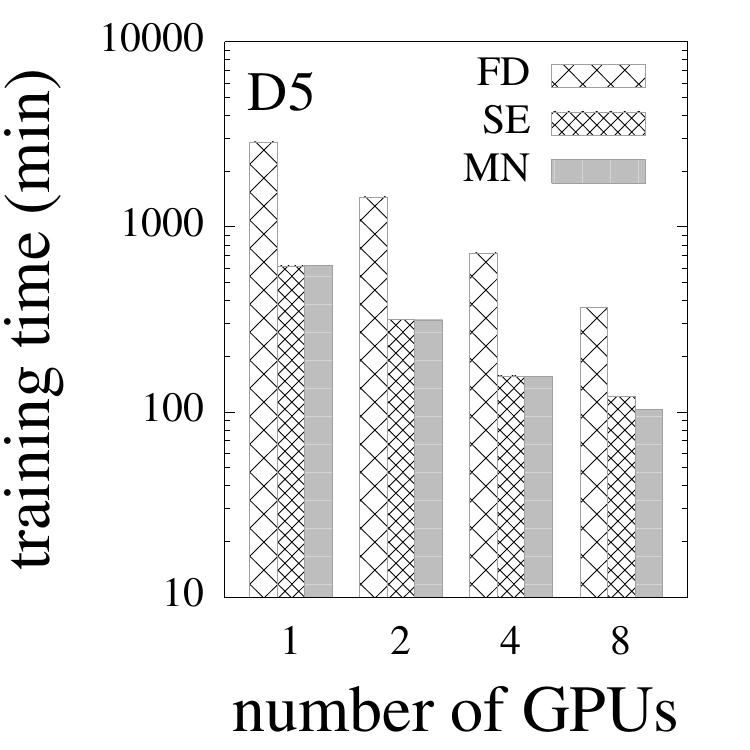}
			\begin{minipage}{0.9\textwidth}
				\caption{\label{exp:parallel_D5}MotherNets 
				is able to utilize multiple GPUs effectively 
				scaling better than SE.}
			\end{minipage}
	\end{minipage}
	\begin{minipage}[t]{0.245\textwidth}
	\includegraphics[width=\textwidth]
		{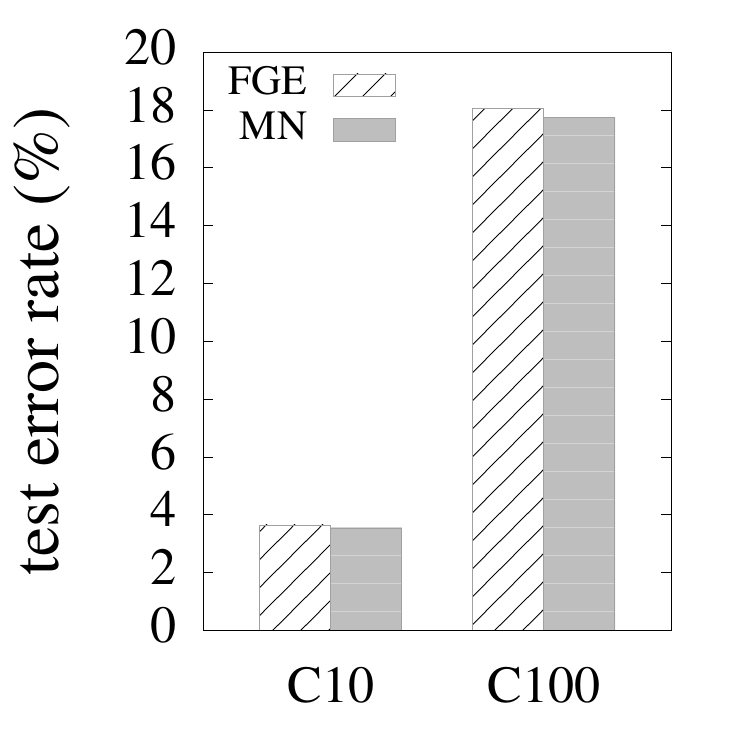}
		\begin{minipage}{0.9\textwidth}
			\caption{\label{exp:FGE_WRN} MotherNets 
			outperform FGE on Wide ResNet 
			ensembles.}
		\end{minipage}
	\end{minipage}
\end{figure*}

\boldPara{Constructing a shared-MotherNet.} 
Given an ensemble $E$ of $K$ hatched networks (i.e., 
those networks that are obtained from a trained 
MotherNet), we construct a shared-MotherNet $S$ as follows: 
First, $S$ is initialized with $K$ input and output layers, one 
for every hatched network. This allows $S$ to produce as many 
as $K$ predictions.  
Then, every hidden layer of $S$ is constructed one-by-one going from 
the input to the output layer and consolidating all neurons 
across all of $E$ that originate from the MotherNet. 
To consolidate a MotherNet neuron at layer $l_i$, we 
first reduce the $k$ copies of that neuron (across all 
$K$ networks in H) to a single copy. 
All inputs to the neuron that may originate from various other 
neurons in the layer $l_{i-1}$ across different hatched networks 
are added together. 
The output of this consolidated neuron is then forwarded to all 
neurons in the next layer $l_{i+1}$ (across all hatched networks) 
which were connected to the consolidated neuron.

Figure \ref{fig:inference} shows an example of how this process 
works for a simple ensemble of three hatched networks. The filled 
circles represent neurons originating from the MotherNet and the 
colored circles represent neurons from ensemble networks. 
To construct the shared-MotherNet (shown on the right), we go 
layer-by-layer consolidating MotherNet neurons. 

The shared-MotherNet is then trained incrementally. 
This proceeds similarly to step 3 from \S 2, however, now through 
the shared-MotherNet, the neurons originating from the MotherNet 
are trained jointly. This results in an ensemble that has $K$ outputs, 
but some parameters between the networks are shared instead of being 
completely independent. This reduces the overall number of parameters, 
improving both the speed and the memory requirement of inference.

\boldPara{Memory reduction.} 
Assume an ensemble $E = \{N_0, N_1, \dots N_{K-1}\}$ of $K$ neural 
networks (where $N_i$ denotes a neural network architecture in the 
ensemble with $|N_i|$ number of parameters) and its MotherNet $M$. 
The number of parameters in the ensemble is reduced by a factor 
of $\chi$ given by: 

\vspace{-0.2in}
$$
\chi = 1 - \frac{k|M|}{\sum_{i=0}^{K-1} |N_i|}
$$

\boldPara{Results.}
Figure \ref{exp:shared_mn} shows how {shared-MotherNets} 
improves inference time for an ensemble of 5 variants of 
VGGNet as described in Table 1. This ensemble is trained on 
the CIFAR-10 data set. We report both overall ensemble test 
error rate and the inference time per image. We see an 
improvement of $2 \times$ with negligible loss in accuracy. 
This improvement is because shared-MotherNets has a reduced 
number of parameters requiring less computation during 
inference time. This improvement scales with the ensemble size.

\label{appendix:shared}

%

\section{Parallel training} 

Deep learning pipelines rely on clusters of multiple 
GPUs to train computationally-intensive neural networks. 
MotherNets continue to improve training time in such cases 
when an ensemble is trained on more than one GPUs. We 
show this experimentally. 

To train an ensemble of multiple networks, we queue all networks 
that are ready to be trained and assign them to available GPUs 
in the following fashion: If the number of ready networks is 
greater than free GPUs, then we assign a separate network to 
every GPU. If the number of ensemble networks available to be 
trained are less than the number of idle GPUs, then we assign 
one network to multiple GPUs dividing idle GPUs equally between 
networks. In such cases, we adopt data parallelism to train a 
network across multiple machines \cite{Dean2012}. 

We train on a cluster of 8 Nvidia K80 GPUs and vary the 
number of available GPUs from 1 to 8. The training hyperparameters 
are the same as described in Section 3. Figure \ref{exp:parallel_V5} 
and Figure \ref{exp:parallel_D5} show the time to train the V5 and 
D5 ensembles respectively across FD, SE, and MotherNets. We observe 
that compared to Snapshot Ensembles, MotherNets (g=1) scale better 
as we increase the number of GPUs. The reason for this is that 
after the MotherNet has been trained, the rest of the ensemble 
networks are all ready to be trained. They can then be trained 
in a way that minimizes communication overhead by assigning 
them to as distinct set of GPUs as possible. 
Snapshot Ensembles, on the other hand, are generated one after 
the other. In a parallel setting this boils down to 
training a single network across multiple GPUs, which incurs 
communication overhead that increases as the number of GPUs 
increases \cite{Keuper2016}.

\label{appendix:parallel}

\section{Improving over Fast Geometric Ensembles}

Now we compare against Fast Geometric Ensembles (FGE), 
a technique closely related to Snapshot Ensembles (SE) 
\cite{Huang2017,Garipov2018}. 
FGE also trains a single neural network architecture and 
saves the network's parameters at various points of its 
training trajectory. 
In particular, FGE uses a cyclical geometric learning rate 
schedule to explore various regions in a neural network's loss 
surface that have a low test error \cite{Garipov2018}. 
As the learning rate reaches its lowest value in a cycle, 
FGE saves `snapshots' of the network parameters. These 
snapshots are then used in an ensemble.  

We compare MotherNets to FGE using an ensemble of Wide 
Residual Networks trained on CIFAR-10 and CIFAR-100 
\cite{Zagoruyko2016}. 
Our experiment consists of an ensemble with six WRN-28-10. 
For MotherNets, we use six variants of this architecture having 
different number of filters and filter widths. For FGE all networks 
in the ensemble are the same as is required by the approach. 
For a fair comparison, the number of parameters are kept 
identical between the two approaches. 
We use the same training hyperparameters as discussed in the 
FGE paper and train for a full training budget of 200 epochs. 
MotherNets is also allocated the same training budget. The 
MotherNet is trained for a $140$ epochs and every ensemble 
network is trained for $10$ epochs after hatching. 
The experimental hardware is the same as outlined in 
Section 3.
Figure \ref{exp:FGE_WRN} shows that for identical training 
budget MotherNets is more accurate than FGE across both data 
sets. 
\label{appendix:FGE}
\balance

\end{document}